\documentclass[10pt,journal,compsoc]{IEEEtran}
\ifCLASSOPTIONcompsoc
  \usepackage[nocompress]{cite}
\else
  \usepackage{cite}
\fi
\ifCLASSINFOpdf
  \usepackage[pdftex]{graphicx}
  \graphicspath{{./figures}}
  \DeclareGraphicsExtensions{.pdf,.jpeg,.png}
\else
\fi
\usepackage{amsmath}
\usepackage{bm}
\usepackage{amssymb}
\usepackage[ruled, noline, longend]{algorithm2e}

\usepackage{array}

\ifCLASSOPTIONcompsoc
  \usepackage[caption=false,font=footnotesize,labelfont=sf,textfont=sf]{subfig}
\else
  \usepackage[caption=false,font=footnotesize]{subfig}
\fi
\usepackage{url}

\usepackage{multirow}
\usepackage{booktabs}
\usepackage{color}
\usepackage{pifont}
\usepackage{changepage}

\newcommand{\bona}{\textit{bonafide} }
\newcommand{\dbname}{HQ-WMCA }

\begin{document}
%
\title {Deep Models and Shortwave Infrared Information to Detect Face Presentation Attacks} 

\author{Guillaume~Heusch,
        Anjith George,
        David Geissb\"uhler,
        Zohreh Mostaani,
        and~S\'ebastien~Marcel,~\IEEEmembership{Senior~Member,~IEEE}
\IEEEcompsocitemizethanks{
\IEEEcompsocthanksitem
Guillaume Heusch, Anjith George, David Geissb\"uhler, Zohreh Mostaani and~S\'ebastien Marcel are with the 
  Idiap Research Institute, Switzerland, e-mails: \{guillaume.heusch, anjith.george, david.geissbuhler, zohreh.mostaani, sebastien.marcel\}@idiap.ch}
}

\IEEEtitleabstractindextext{%
\begin{abstract}
  This paper addresses the problem of face presentation attack detection using different image modalities. In particular, the usage
of short wave infrared (SWIR) imaging is considered. 
  Face presentation attack detection is performed using recent models based on Convolutional Neural Networks 
  using only carefully selected SWIR image differences as input. Conducted experiments show superior performance over similar models
acting on either color images or on a combination of different modalities (visible, NIR, thermal and depth), as well as on a SVM-based classifier 
acting on SWIR image differences. Experiments have been carried on a new public and freely available database, 
containing a wide variety of attacks. 
Video sequences have been recorded thanks to several
sensors resulting in 14 different streams in the visible, NIR, SWIR and thermal spectra, as well as depth data.
The best proposed approach is able to almost perfectly detect all impersonation attacks while ensuring low \bona classification
errors. On the other hand, obtained results show that obfuscation attacks are more difficult to detect. We hope that the proposed
database will foster research on this challenging problem. 
Finally, all the code and instructions to reproduce presented
experiments is made available to the research community.

\end{abstract}

}

\maketitle

\IEEEdisplaynontitleabstractindextext

%
\IEEEpeerreviewmaketitle

\IEEEraisesectionheading{\section{Introduction}\label{sec:introduction}}

%
%
%
%



Biometrics is nowadays used in a variety of scenarios and is becoming
a standard mean for identity verification. Among the different modalities, 
face is certainly the most used, since it is both convenient and, in most cases, sufficiently reliable.
Nevertheless, there exists many studies showing that current face recognition 
algorithms are not robust to face presentation attacks
\cite{kose-icassp-2013} \cite{hadid-cvprw-2014} \cite{chingovska-frais-2016} \cite{mohammadi-iet-2017} \cite{bhattacharjee-btas-2018}. 
A presentation attack consists in presenting a fake (or altered) biometric sample to 
a sensor in order to fool it. For instance, a fingerprint reader can be tricked by 
a fake finger made of playdough. For the face modality, examples of attacks range from a simple photograph to more sophisticated silicone masks.
For a wide acceptance of the face biometric
as an identity verification mean, face recognition systems should be robust to presentation attacks. Consequently, 
numerous presentation attack detection (PAD) approaches have been proposed in the last decade, and surveys can be found in \cite{galbally-access-2014} and \cite{li-iet-2018}.
Existing PAD algorithms are usually classified based on the information they act upon. Some rely on liveness information, such as blinking eyes \cite{pan-iccv-2007} or 
blood pulse information \cite{heusch-btas-2018}. Others take advantage of the differences between \bona attempts and 
attacks through the use of texture \cite{chingovska-biosig-2012}, image quality measures \cite{wen-tifs-2015} or frequency analysis \cite{caetano-tifs-2015}. 
As expected, there also exists approaches relying on deep Convolutional Neural Networks (CNN): relevant examples can be found in \cite{yang-arxiv-2014} and 
\cite{li-tifs-2018-2}.\\

While most of the literature presents PAD algorithms acting on traditional RGB data, some works also suggest to tackle presentation
attacks using images from different modalities \cite{george-tifs-2019,george-tifs-2020,nikisins2019domain,george2020face}. For instance, depth information has been used in conjunction 
with color images in \cite{atoum-ijcb-2017}. Yi \textit{et al.} \cite{yi-hbas-2014} combines the visible and near infrared
(NIR) spectrum to improve robustness against photo attacks. Thermal imaging has also been investigated to detect mask attacks in \cite{bhattacharjee-biosig-2017}.
Steiner \textit{et al.} \cite{steiner-icb-2016} proposed an approach based on short-wave infrared (SWIR) images to discriminate
skin from non-skin pixels in face images. Also, processing data from different domains with CNNs has been successfully applied to presentation attack detection: 
For instance, Tolosana \textit{et al.} \cite{tolosana-tifs-2019} used SWIR imaging in conjunction with classical deep models to detect fake fingers. 
Regarding face PAD, George \textit{et al.} \cite{george-tifs-2019} proposed a multi-channel CNN combining visual, NIR, depth and thermal information. Authors showed that 
this model can achieve a very low error rate on a wide variety of attacks, including printed photographs, video replays and a variety of masks. 
Parkin and Grinchuk \cite{parkin-cvprw-2019} also recently proposed a multi-channel face PAD network with different ResNet blocks for different channels. 
Before fusing the channels, squeeze and excitation modules are used, followed by additional residual blocks. 
Furthermore, aggregation blocks at multiple levels are added to leverage inter-channel correlations. Their final PAD method
averages the output over 24 such models, each trained with different settings (i.e. on different attack types for instance). 
It achieved state-of-the-art performance on the CASIA-SURF database \cite{zhang-arxiv-2018}, where only print attacks are considered. \\

Among the different used sensors, SWIR imaging seems promising. Indeed, one of its main features is that water is very absorbing in some SWIR wavelengths. 
For instance, SWIR imaging is used for food inspection and sorting, based on water content \cite{xenics}.
Since 50 to 75 \% of the human body is made of water, this modality is hence very relevant for face PAD.  While SWIR imaging has already been 
studied in the context of face recognition \cite{bourlai-icpr-2010} \cite{nicolo-tifs-2012}, there are very few works on this modality in the context of face PAD.
Actually, at the time of writing, there is only one such study made by Steiner and colleagues \cite{steiner-icb-2016}. 
This is arguably due to the lack of available data: the only database containing face presentation attack in SWIR is the BRSU database, introduced in \cite{steiner-sensors-2015}.
The BRSU database contains \bona images of 53 subjects (there are 3 to 4 frontal face images per subject) and 84 images of various attacks performed by 5 subjects. 
While comprising a relatively large diversity in terms of attack types (masks, makeup and various disguises), this database is quite small. It is hence not suited 
to assess latest approaches in face PAD leveraging CNNs. Furthermore, images in the visible spectrum and at various SWIR wavelengths are not aligned, making face registration 
more difficult.\\

In this contribution, the usage of CNNs in conjunction with SWIR information is investigated to address face presentation attack detection. 
Two recent models for face PAD are considered: the Multi-Channel CNN proposed in \cite{george-tifs-2019} and a multi-channel extension of the network with Pixel-wise Binary Supervision 
proposed in \cite{george-icb-2019}. These approaches were selected for their capacity to handle multimodal data, 
their simplicity (i.e. the training procedures are straightforward) and their good performance when different attack types are considered.
These models are fed with a combination of SWIR image differences, which have been selected using a sequential feature selection algorithm.
To assess the effectiveness of the proposed approach, a new publicly and freely available database, \dbname, is introduced. 
It contains video sequences of both \bona authentication attempts and attacks recorded with co-registered RGB, depth and multispectral (NIR, SWIR and thermal) sensors. 
Moreover, it contains many presentation attack instruments (PAIs) including disguise (tattoos, make-up and wigs) alongside more traditional attacks, such as photographs, 
replays and a variety of masks.\\

The rest of the paper is organized as follows. Section \ref{sec:pad} presents the 
transformation and the selection process applied on recorded SWIR data and the two investigated CNNs in more details. 
Section \ref{sec:database} introduces the new \dbname database: in particular, it presents the hardware setup, the different PAIs and the
experimental protocols. Section \ref{sec:experiments} is devoted to the experimental evaluation. After having introduced the experimental framework, the two models using SWIR data
are evaluated on the proposed database and compared to different baselines, including an SVM-based classifier acting on SWIR data and previously proposed CNNs
using other image modalities. Finally, Section \ref{sec:conclusion} concludes the paper.

\section{Presentation Attack Detection Approach}
\label{sec:pad}
In this section, the approaches to face presentation attack detection are presented. The usage of
SWIR data is explained before proceeding with the description of the two convolutional neural
networks that were considered.

\subsection{SWIR data}
\label{sec:pad-swir}

\begin{figure}[h!]
\centering
  \subfloat{\includegraphics[height=2.5cm]{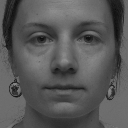}}%
\hfil
  \subfloat{\includegraphics[height=2.5cm]{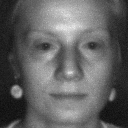}}%
\hfil
  \subfloat{\includegraphics[height=2.5cm]{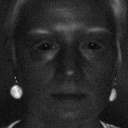}}%
  \\
  \subfloat{\includegraphics[height=2.5cm]{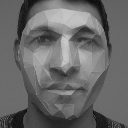}}%
\hfil
  \subfloat{\includegraphics[height=2.5cm]{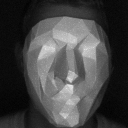}}%
\hfil
  \subfloat{\includegraphics[height=2.5cm]{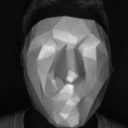}}%
  \caption{A \bona face image and a paper mask attack in the visible spectrum (left), at a wavelength of 940nm band (center) and at a wavelength of 1450nm band (right). 
  Note how the skin appears darker at 1450nm on the \bona face image: this
  is due to water absorption. On the other hand, this phenomenon is not happening with the mask attack.}
\label{fig:vis-vs-swirs}
\end{figure}

As mentioned in the introduction, SWIR data has not (yet) been widely used in face-related tasks, despite its
interesting properties. It has been shown in \cite{wilson-jbo-2015} that, for water, absorption
peaks near 1430nm and this behavior is particularly suitable for detecting non-skin material. 
Indeed, this was already mentioned in \cite{nicolo-tifs-2012}: 'The human skin and eyes in the 
SWIR spectrum appear to be very dark because of the presence of moisture'. This phenomenon is illustrated in Figure~\ref{fig:vis-vs-swirs}, where
the face image of a \bona attempt and of a paper mask attack are shown in different part of the spectrum.\\

\subsubsection{Normalized difference}

Instead of directly using images at different SWIR wavelengths, a normalized difference between these images has been considered, as done
in both \cite{steiner-sensors-2015} and \cite{tolosana-tifs-2019}. This normalization is independent of the absolute brightness and
exhibits differences between skin and non-skin pixels \cite{steiner-sensors-2015}.
Consider two SWIR images of the same individual, $I_{s_1}$ and $I_{s_2}$, recorded at (almost) the same time\footnote{There is a lag of 11ms between frames recorded at different wavelength, 
resulting in a total lag of 77ms within the considered SWIR range.} but at different wavelengths, the normalized difference is given by:

\begin{equation}
\label{eq:swir-diff}
  d(I_{s_1}, I_{s_2}) = \frac{I_{s_1} - I_{s_2}}{I_{s_1} + I_{s_2} + \epsilon}
\end{equation}

In our work, $\epsilon$ was set to $1e^{-4}$. For some reason, previous works \cite{steiner-sensors-2015} \cite{tolosana-tifs-2019} 
only consider differences between $n$ SWIR bands with $1 \leq s_1 < n-1$ and $s_1 < s_2 < n$. 
However, the subtraction operation is not commutative, i.e. $d(s_1, s_2) \neq d(s_2 ,s_1)$, hence in our work, all differences are considered.
This allows to have more input data, with possible complementary information, as opposed to previous approaches.
Since our recording setup allows to capture SWIR data at no less than $n=7$ different wavelength in each recordings, the number of possible 
SWIR image differences is hence given by:

\begin{equation}
\frac{n!}{(n-2)!} = \frac{7!}{5!} = 6 \cdot 7 = 42
\end{equation}

These $42$ differential images may surely be correlated, and only a particular subset can contain information relevant to face PAD. 
Furthermore, some of these images may not contain any relevant information at all. 
Consequently, particular care should be made in the selection of the most useful subset of such differences. 
The procedure to perform such a selection is explained in more details below.

\subsubsection{SWIR Images Differences Selection}
Consider the set containing the 42 possible differences: $S = \{d(I_{s_1}, I_{s_2}), ..., d(I_{s_7}, I_{s_6})\}$. As
a first step, and similar to \cite{tolosana-tifs-2019},
the set has first been ordered according to the inter-class to intra-class variability ratio, computed in terms of pixel-wise difference. 
The pseudo-code for the algorithm
to sort the set of difference is presented in Algorithm~\ref{algo:intra-inter}. For the sake of clarity, an example $e^i$ is considered to consist in 
7 SWIR images (and not video sequences) at different wavelengths. Note also that the division in the penultimate line of Algorithm~\ref{algo:intra-inter} is done
element-wise on the 42-dimensional vectors containing the mean inter and intra-class distances. 
At the end of the procedure, the pixel-wise inter/intra-class ratio is obtained for each of the 42 differences.
The ordered set is then given by sorting the 42 ratios, beginning with the highest.

\SetKwInput{KwInput}{Input}           
\SetKwInput{KwOutput}{Output} 
\SetKwInput{KwInit}{Initialization}
\SetKwIF{If}{ElseIf}{Else}{If}{:}{elif}{else:}{}%

\begin{algorithm}
\label{algo:intra-inter}
\DontPrintSemicolon

  \KwInput{$E = \{e^1, e^2, ..., e^n\}$: set of examples}
  \KwInit{$k_{bf}, k_{a} = 0$, intra = $\mathbf{0}$, inter = $\mathbf{0}$ }
  \KwOutput{$S^*$: ordered set of SWIR differences}
  
  \For{$e^i, e^j \in E$ $\forall i, j, i \neq j$}
  {
    $S_i = [mean(d(e^i_{s_1}, e^i_{s_2})), ..., mean(d(e^i_{s_7}, e^i_{s_6}))]$\;
    $S_j = [mean(d(e^j_{s_1}, e^j_{s_2})), ..., mean(d(e^j_{s_7}, e^j_{s_6}))]$\;
    $\Delta_{i,j} = |S_i - S_j|$\;

    \If{$e^i$ and $e^j$ are \bona}
    {
      intra = intra + $\Delta_{i,j}$,
      $k_{bf} = k_{bf} + 1$

    }
    \If{$e^i$ is \bona and $e^j$ is attack}
    {
      inter = inter + $\Delta_{i,j}$,
      $k_{a} = k_{a} + 1$
    }
    \If{$e^i$ is attack and $e^j$ is \bona}
    {
      inter = inter + $\Delta_{i,j}$,
      $k_{a} = k_{a} + 1$
    }
  }
  intra = intra / $k_{bf}$, 
  inter = inter / $k_{a}$\;
  ratio = inter / intra\;
  $S^*$ = sort(ratio)

\caption{Pixel-wise intra/inter class ratio.}
\end{algorithm}

For this purpose, only the training set of the \dbname database has been used. This gives a first insight 
on the discriminative power of each of the differences between different SWIR bands for face PAD. 
In \cite{tolosana-tifs-2019} the 3 most informative SWIR image differences, according
to this criterion, are used to feed CNNs. In our work, it is proposed to extend this approach by subsequently applying 
a mechanism to automatically select the best subset of such differences. As opposed to \cite{tolosana-tifs-2019}, our approach
is taking the task at hand into account.
For this purpose, a sequential forward floating selection
(SFFS) mechanism \cite{pudil-prl-1994} has been applied on the ordered set to select the optimal subset of SWIR differences.
The criterion $\mathcal{J}$ used here is the average classification error rate (ACER) on the development set of the database.
Basically, the SFFS algorithm will sequentially add features (i.e. SWIR image differences) as input to the CNNs model, and retain
the ones which improves performance. Each time a feature is retained, a "backward" step is performed to check 
if removing a particular input feature further improves. The SFFS algorithm is presented in Algorithm~\ref{algo:sffs}.

\SetKwInput{KwInput}{Input}           
\SetKwInput{KwOutput}{Output} 
\SetKwInput{KwInit}{Initialization} 
\begin{algorithm}
\label{algo:sffs}
\DontPrintSemicolon
  
  \KwInput{$\{s_1, s_2, ..., s_n\}$: ordered set of SWIR differences}
  \KwInit{$e^* = 100.0, S^* = \emptyset$}
  \KwOutput{$S^*, e^*$} 

  \For{$i = 1$ \KwTo $n$}
  {
    $S = S^* \cup s_i$ \;
    $e = \mathcal{J}(S)$ \;
    \If{$e < e^*$}
    {
      $S^* = S^* \cup s_i$ \;
      $e^* = e$ \;
      
      \For{$j = 1$ \KwTo $|S^*| - 1$, $j \neq i$}
      {
        $S = S^* \setminus s_j$ \;
        $e = \mathcal{J}(S)$ \;
        \If{$e < e^*$}
        {
          $S^* = S^* \setminus s_j$ \;
          $e^* = e$ \;
        }
      }
    }
  }
\caption{Sequential Forward Floating Selection}
\end{algorithm}

\subsection{Deep Convolutional Networks}

Deep CNN-based PAD methods have consistently outperformed feature based methods, 
which holds true in a multimodal setting as well \cite{george-tifs-2019}. 
In this work two different models were used, corresponding to early fusion and late fusion strategies. 
The main idea is to leverage the joint representation from information coming from different sources to reliably detect presentation attacks. 
Note that both investigated models use a specific backbone architecture (LightCNN for MC-CNN and DenseNet for MC-PixBiS). 
While other, more recent backbones can be used within these frameworks, 
it has been decided to stick with backbones proposed in their original implementation for the sake of comparison.
The two architectures are presented in more details below. 

\subsubsection{Multi-Channel CNN}

\begin{figure}[ht]
\centering
\includegraphics[width=1\columnwidth]{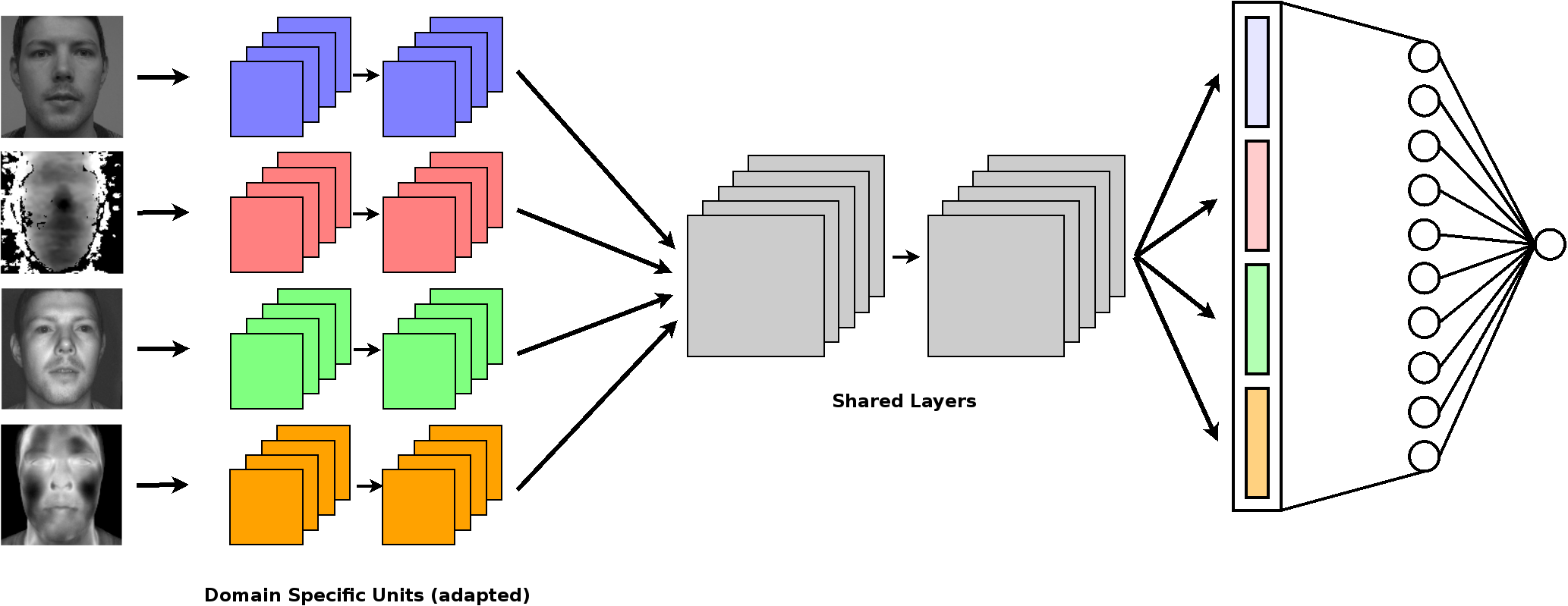}
  \caption{Block diagram of the MC-CNN network. The gray color blocks in the CNN part represent layers which are not retrained, and other colored blocks represent re-trained/adapted layers.
  Note that the original approach from \cite{george-tifs-2019} is depicted here: it takes grayscale, infrared, depth and thermal data as input. In our work, these inputs are replaced
  with a variable number of SWIR images differences.}
\label{fig:mccnn}
\end{figure}

The main idea in the Multi-Channel CNN (MC-CNN) is to use the joint representation from multiple modalities for PAD, using transfer learning from a pre-trained face recognition network \cite{george-tifs-2019}. 
The underlying hypothesis is that the joint representation in the face space could contain discriminative information for PAD. This network consists of three parts: 
low and high level convolutional/pooling layers, and fully connected layers, as shown in Figure~\ref{fig:mccnn}.
As noted in \cite{pereira-tifs-2019}, high-level features in deep convolutional neural networks trained in the visual spectrum are domain independent i.e. they do not depend on a specific modality. 
Consequently, they can be used to encode face images collected from different image sensing domains. 
The parameters of this CNN can then be split into higher level layers (shared among the different channels), and lower level layers (known as Domain Specific Units).
By concatenating the representation from different channels and using fully connected layers, a decision boundary for the appearance of \bona and attack presentations can be learned via back-propagation. 
During training, low level layers are adapted separately for different modalities, while shared higher level layers remain unaltered.
In the last part of the network, embeddings extracted from all modalities are concatenated, and two fully connected layers are added. 
The first fully connected layer has ten nodes, and the second one has one node. Sigmoidal activation functions are used in each fully connected layer, as in the original implementation \cite{george-tifs-2019}. 
These layers, added on top of the concatenated representations, are tuned exclusively for the PAD task using the Binary Cross Entropy as the loss function.
The MC-CNN approach hence introduces a novel solution for multimodal PAD problems, leveraging a pre-trained network for face recognition 
when a limited amount of data is available for training PAD systems. Note that this architecture can be easily extended for an arbitrary number of input channels.

\subsubsection{Multi-Channel Deep Pixel-wise Binary Supervision}

\begin{figure*}
\centering
\includegraphics[width=0.95\textwidth]{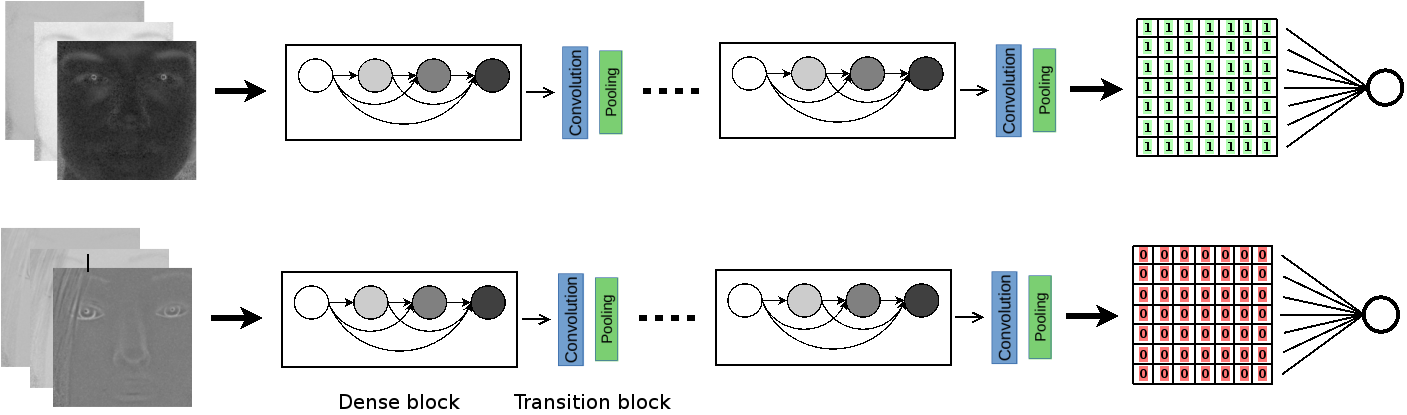}
  \caption{MC-PixBiS architecture with pixel-wise supervision. SWIR images differences are stacked before being passed to a series of dense blocks. 
  A 1x1 convolution is then applied to yield the 14x14 supervision map. The top row shows the network fed with a \bona example, and consequently, the ground truth for the supervision map 
  is composed of ones. The bottom row shows the same network fed with an attack: in this case, the ground truth consists in zeros. The supervision map is used to compute 
  the first part of the loss $\lambda \mathcal{L}_{pix}$ in Equation~\ref{eq:pixbis-loss}. Finally, the supervision map is flattened and a fed to a linear layer with sigmoid activation. The final node is a binary output
  representing the probability of the presented example to be \bona, and is used to compute $\mathcal{L}_{bin}$.} 
  
\label{fig:pixbis}
\end{figure*}

The Multi-Channel Deep Pixel-wise Binary Supervision network (MC-PixBiS) is a multi-channel extension of a recently published work on face PAD using legacy RGB sensors \cite{george-icb-2019}. 
The main idea in \cite{george-icb-2019} is to use pixel-wise supervision as an auxiliary supervision. 
The pixel-wise supervision forces the network to learn shared representations, and it acts like a patch wise method (see Figure~\ref{fig:pixbis}). 
To extend this network for a multimodal scenario, we use the method proposed in \cite{wang-eccv-2016}: averaging the filters in the first layer and replicating the weights for different modalities.

The general block diagram of the framework is shown in Figure \ref{fig:pixbis} and is based on DenseNet \cite{huang-cvpr-2017}. The first part of the network contains eight layers, and each layer
consists of two dense blocks and two transition blocks. The dense blocks consist of dense connections between every layer with the same feature map size, and
the transition blocks normalize and downsample the feature maps. The output from the eighth layer is a map of size $14\times14$ with 384 features. 
A $1 \times 1$ convolution layer is added along with sigmoid activation to produce the binary feature map. 
Further, a fully connected layer with sigmoid activation is added to produce the binary output. A combination of  
losses is used as the objective function to minimize: 
\begin{equation}
\label{eq:pixbis-loss}
\mathcal{L} = \lambda \mathcal{L}_{pix}+ (1-\lambda)\mathcal{L}_{bin}
\end{equation}

where $\mathcal{L}_{pix}$ is the binary cross-entropy loss applied to each element of the $14 \times 14$ binary output map and $\mathcal{L}_{bin}$ is
the binary cross-entropy loss on the network's binary output.
A $\lambda$ value of 0.5 was used in our implementation. Even though both losses are used in training, in the evaluation phase, only the pixel-wise map is used: 
the mean value of the generated map is used as a score reflecting the probability of \bona presentation.

\section{The \dbname Database}
\label{sec:database}
In this section, the new High-Quality Wide Multi-Channel Attack database, \dbname is described. 
This database can be viewed as an extension of the WMCA database previously
presented in \cite{george-tifs-2019}. The proposed database is however different in  
several important aspects. Firstly, the various sensors used to capture data are of
better quality and hence allowed to record video sequences at a higher resolution and 
at a higher frame rate than for the WMCA database. Furthermore,  a new sensor 
acting in the shortwave infrared (SWIR) spectrum has been added. Additionally, and thanks
to a dedicated illumination module, several NIR and SWIR wavelengths have been captured. 
Secondly, the proposed database contains a wider range of attacks. In particular, it incorporates
obfuscation attacks, where the attacker tries to hide its identity. In the remainder of this Section, 
the hardware setup and sensors characteristics are first presented. The procedure for data recording and
a description of the different attacks is then made before proceeding with
the experimental protocols.

\subsection{Hardware Setup}

\begin{figure}[h]
\centering
\includegraphics[width=0.4\textwidth]{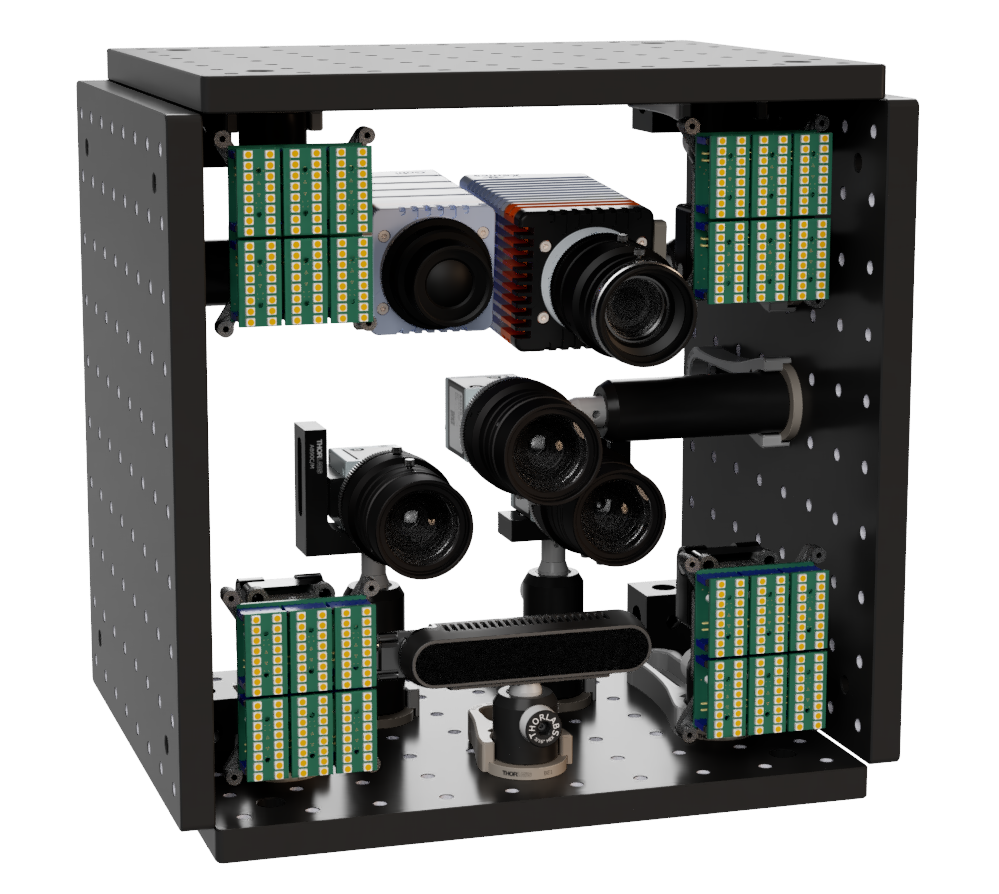}
\caption{Face biometric sensor suite.}
\label{fig:face-station}
\end{figure}

Data were recorded thanks to a custom made sensors suite with several cameras,
as shown in Figure~\ref{fig:face-station}. 
These sensors allowed to record
both genuine faces and presentation attacks in no less than five different image modalities: RGB, NIR, SWIR, thermal
and depth. Information about the different sensors can be found in Table~\ref{tab:sensors}.

\begin{table}[h]
\centering
\caption{Sensors description}
\label{tab:sensors}
\begin{tabular}{llcc}
\toprule
Sensor name            & Modality & Resolution       & Frame rate \\ 
\midrule
Basler acA1921-150uc   & Color   & 1920$\times$1200 & 30\\
Basler acA1920-150um   & NIR     & 1920$\times$1200 & 90\\
Xenics Bobcat-640-GigE & SWIR    & 640$\times$512   & 90\\
Xenics Gobi-640-GigE   & Thermal & 640$\times$480   & 30\\
Intel Realsense D415   & Depth   & 720$\times$1280   & 30\\
\bottomrule
\end{tabular}
\end{table}

Furthermore, four banks of 6 Light Emitting Diodes (LED) modules are used for illumination besides the ambient illumination available in the room. 
Each LED module consists of LEDs operating in 10 different wavelengths from 735nm to 1650nm, covering NIR and SWIR spectra. 
Sequencial switching of these infrared emmitters, synchronized with cameras exposure periods, therefore yields a measure of multi-spectral reflectivity across the sample. 
These wavelenghts were selected to give the best possible multi-spectral coverage given market availability.
As a result, each recording contains data in 14 different "modalities", including 4 NIR and 7 SWIR wavelengths.
All cameras have been co-registered thanks to a calibration procedure, allowing the captured data to be aligned in each of the modalities. 

\subsection{Data collection procedure}
The data collection was performed during three sessions, which were typically recorded one week apart.  
The sessions were different based on their illumination environment. The first session was recorded with ceiling office light, 
the second using an additional halogen lamp, and the third one only with LED spotlights facing the subject, and without any other light source.
During each session, data for \bona and at least three presentation attacks performed by the participants were captured, 
as well as some of the presentation attacks presented on a stand. Since the duration of a recording was only 2 seconds, 
it was repeated twice to include more samples.

The participants were asked to sit in front of the cameras and look towards the sensors with a neutral facial expression. 
The sensors were located at a distance of 50-60 cm for both \bona and presentation attacks. 
If the subjects wore medical glasses, their \bona data were captured twice, with and without glasses. 
Some of the presentation attacks such as masks and mannequins were heated up before the data capture. This was done in order
to reach a temperature close to the one of human body, to avoid a too easy detection of such attacks with the thermal sensor. 
The acquisition operator made sure that the face was visible in all the sensors before recording.

\begin{figure*}[t]
\centering
  \subfloat[]{\includegraphics[height=2.5cm]{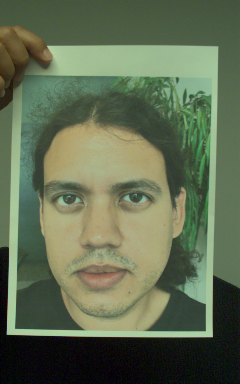}}%
\hfil
  \subfloat[]{\includegraphics[height=2.5cm]{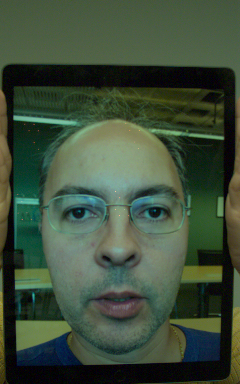}}%
\hfil
  \subfloat[]{\includegraphics[height=2.5cm]{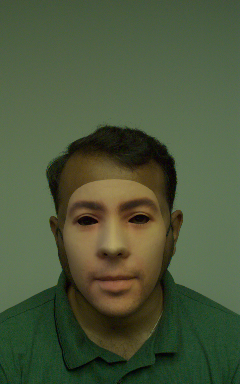}}%
\hfil
  \subfloat[]{\includegraphics[height=2.5cm]{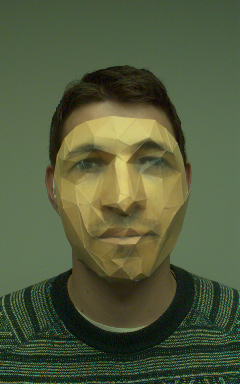}}%
\hfil
  \subfloat[]{\includegraphics[height=2.5cm]{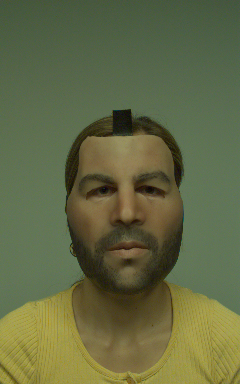}}%
\hfil
  \subfloat[]{\includegraphics[height=2.5cm]{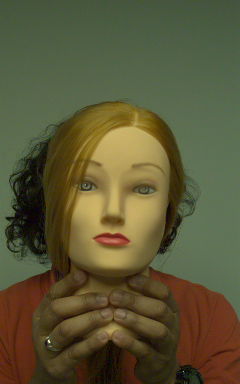}}%
\hfil
  \subfloat[]{\includegraphics[height=2.5cm]{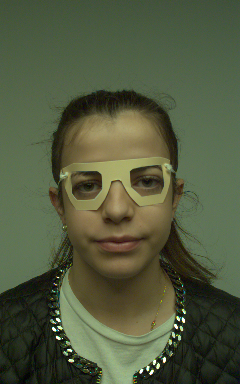}}%
\hfil
  \subfloat[]{\includegraphics[height=2.5cm]{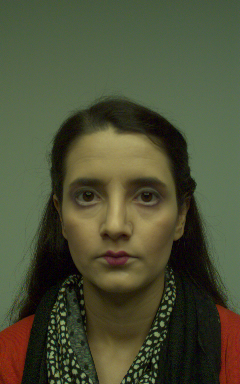}}%
\hfil
  \subfloat[]{\includegraphics[height=2.5cm]{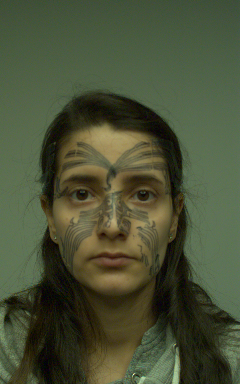}}%
\hfil
  \subfloat[]{\includegraphics[height=2.5cm]{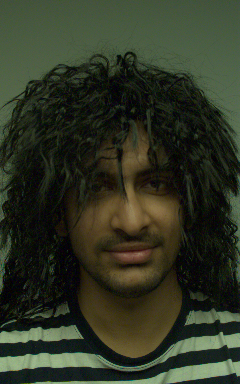}}%
  \caption{Example of attacks present in the database. (a) Print, (b) Replay, (c) Rigid mask, (d) Paper mask, (e) Flexible mask, (f) Mannequin, 
          (g) Glasses, (h) Makeup, (i) Tattoo and (j) Wig.
          Note that only one particular example for each category is shown here, but there exists more variation across the database. 
          For instance, print attacks have been crafted using different printers and different papers.}
\label{fig:attacks}
\end{figure*}

The presentation attacks in the database have been captured by presenting more than 100 different Presentation Attack Instruments (PAIs) to the cameras. 
The PAIs can be grouped into ten different categories, as listed below. An example for each category is shown in Figure~\ref{fig:attacks}. 
Note also that no attack combination (i.e. glasses and makeup at the same time) have been considered.

\begin{itemize}
    \item \textbf{Glasses:} A clear lens glasses with a large frame, different models of decorative glasses with printed eyes, and paper glasses.
    \item \textbf{Mannequin:} Several models of mannequin heads.
    \item \textbf{Print}: Printed photograph of faces on either matte or glossy paper using a laser printer (CX c224e) and an  
                          inkjet printer (Epson XP-860). The original photos were resized so that the size of the printed face is within the range of a human face.
    \item \textbf{Replay:} Videos while played or paused and digital photos presented on an iPad Pro 12.9in. The original videos used to perform presentation
                           attacks were captured in HD at 30 fps by the front camera of an iPhone 6S and in full-HD at 30 fps by the rear camera of the iPad Pro. 
                           Some of the videos were resized so that the size of the face presented on the display is human like and therefore their quality vary.
    \item \textbf{Rigid mask:} Different types of plastic masks: non-transparent, transparent without makeup, and transparent with makeup, and custom made realistic rigid masks.
    \item \textbf{Flexible mask:} Custom made realistic soft silicon masks.
    \item \textbf{Paper mask:} Custom paper masks were made by printing photos of real identities on matte and glossy papers using printers mentioned in the "Print" category.
    \item \textbf{Wigs:} Several models of wigs for men and women.
    \item \textbf{Tattoo:} Removable facial tattoos based on Maori tribal face tattoo.
    \item \textbf{Makeup:} Three different methods of makeup were performed during the data collection namely ``Heavy Contour'', ``Pattern'', and ``Transformation''. 
                           The first two methods were performed in three levels of intensity and were designed to change the shape of the contours and the regular shadows of the face. 
                           The last method was used to transform the face of the participant to impersonate another identity, normally a famous character. 
                           The data for the latter method was only captured with one level of intensity and in order to compensate for the lack of data in this case, 
                           such makeup attacks were captured three times as opposed to two times for other presentations.  
\end{itemize}

There is a total of 2904 multi-modal presentation video sequences, for a total of 58080 images (in each modality) in the database: 555 \bona presentations from 51 participants and the remaining 2349 are presentation attacks. 
This database is made freely and publicly available for research purposes\footnote{\url{https://www.idiap.ch/dataset/hq-wmca}}. 

\subsection{Experimental Protocols}
\label{sec:protocols}
As it is standard practice with classification problems using machine learning, data has been divided into
three sets: train, validation and test. For an unbiased evaluation, there is no overlap in identities of the \bona examples among the different 
sets. The statistics for each set are given in Table~\ref{tab:set-stats}. As can be seen on Table~\ref{tab:attacks-distribution}, special care has 
been taken to balance attacks across the different sets. Note finally that each example consists of 10 frames, evenly sampled along the video sequence. 

\begin{table}[h!]
  \centering
  \caption{Number of examples for \bona and attack examples in each set. The number of different identities
  is given in parenthesis. Note that while having different identities provides variability for \bona examples, 
  the total number of identities is not critical to assess the performance of PAD. Rather, the number and the variability 
  in the attacks should be considered.}
  \begin{tabular}{llll}
    & \textbf{Train} & \textbf{Validation} & \textbf{Test}\\
  \midrule
    \textit{Bonafide} & 228 (21) & 145 (14) & 182 (16)\\
    Attacks & 742 & 823 & 784\\
  \bottomrule
  \end{tabular}
  \label{tab:set-stats}
\end{table}

\begin{table}
  \centering
  \caption{Distribution of attacks in the different sets}
  \begin{tabular}{llccc}
    & \textbf{Attack type} & \textbf{Train} & \textbf{Validation} & \textbf{Test}\\
  \midrule
    \multirow{7}{*}{\textbf{Impersonation}} & Print         & 48  & 98  & 0\\
                                            & Replay        & 36  & 100 & 126\\
                                            & Rigid Mask    & 162 & 118 & 140\\
                                            & Paper Mask    & 28  & 24  & 49\\
                                            & Flexible Mask & 90  & 86  & 48\\
                                            & Mannequin     & 20  & 38  & 77\\
                                            & \it{Total}    & \it{384} & \it{464} & \it{440}\\
  \midrule
    \multirow{5}{*}{\textbf{Obfuscation}} & Glasses     & 56  & 38  & 36\\
                                          & Makeup      & 264 & 271 & 258\\
                                          & Tattoo      & 24  & 24  & 24\\
                                          & Wig         & 14  & 26  & 26\\
                                          & \it{Total}  & \it{358} & \it{359} & \it{344}\\
  \bottomrule
  \end{tabular}
  \label{tab:attacks-distribution}
\end{table}

In this contribution, different experimental scenarios are considered. Indeed, experiments
have been performed in three different settings:

\begin{enumerate}
  \item\textbf{Grand Test:}
This scenario considers all possible attacks, and thus allows to assess the ability of different 
PAD approaches to handle a wide variety of attacks.
  
  \item\textbf{Impersonation attacks:}
Impersonation attacks are defined as attacks in which the attacker tries to 
authenticate himself or herself as another person. Attacks corresponding to 
this scenario are prints, replays and masks. Note that masks are not 
necessarily representing a real, existing identity. In this work however, all 
mask attacks are considered as impersonation attacks since they usually cover the whole
face of the attacker. This protocol has been implemented by removing all obfuscation attacks
present in the grand test scenario.

  \item\textbf{Obfuscation attacks:}
In the case of obfuscation attacks, the appearance of the attacker is altered in the 
hope of not being properly recognized by a face recognition system. Attacks corresponding
to this scenario are typically various forms of disguises, such as glasses, wigs, makeup
and tattoos. This protocol has been implemented by removing all impersonation attacks
present in the grand test scenario. Note that it is debatable whether such examples should be
    considered as attacks \textit{per se}, since the person does not necessarily try to bypass a face recognition
system by being identified as someone else. Nevertheless, the ISO/IEC 30107-3 standard \cite{ISO-30107-3} defines such 
    \textit{concealer} attacks as a possible mean to defeat any given face recognition system.
    Besides, several studies adressed such disguises 
    in the context of face recognition, such as \cite{chen-isba-2017}, \cite{singh-tbiom-2019} and more recently \cite{deng-iccvw-2019}, which
    won the Disguised Faces in the Wild challenge. 
It has been consequently decided to consider such attacks, since they actually impair the correct
    operation of a face recognition system: it is thus important to detect them.

\end{enumerate}

\section{Experiments \& Results}
\label{sec:experiments}
In this section, the performance measures and the experimental setup are first presented.
Then, results for different baselines and proposed approaches are presented and discussed.

\subsection{Performance Measures}
Any face presentation attack detection algorithm encounters two types of error: either \bona
attempts are wrongly classified as attacks, or the other way around, i.e. an attack is misclassified as a real attempt.
As a consequence, performance is usually assessed using two metrics. 
The Attack Presentation Classification Error Rate (APCER) is defined as the expected probability of a 
successful attack and is defined as follows:
\begin{equation}
  APCER = \frac{\text{ \# of accepted attacks}}{\text{ \# of attacks}}
\end{equation}
Conversely, the Bonafide Presentation Classification Error Rate (BPCER) is defined as the expected probability 
that a \bona attempt will be falsely declared as a presentation attack. The BPCER is computed as:
\begin{equation}
  BPCER = \frac{\text{ \# of rejected real attempts}}{\text{ \# of real attempts}}
\end{equation}
Note that according to the ISO/IEC 30107-3 standard \cite{ISO-30107-3}, each attack type should be taken into account separately. 
We did not follow this standard here, since our goal is to assess the robustness for a wide range of attacks. 
To provide a single number for the performance, results are typically presented using the
Average Classification Error Rate (ACER), which is basically the mean of the APCER and the BPCER:
\begin{equation}
\label{eq:ACER}
  ACER(\tau) = \frac{APCER(\tau) + BPCER(\tau)}{2} \quad \textrm{[\%]} 
\end{equation}                                                                                                                                              
Note that the ACER depends on a threshold $\tau$. Indeed, 
reducing the APCER will increase the BPCER and vice-versa. 
For this reason, results are often presented using either Receiver Operating Characteristic (ROC) 
or Detection-Error Tradeoff (DET) curves, which
plot the  APCER versus the BPCER for different thresholds~\cite{martin-eurospeech-1997}.
In our work, the APCER at BPCER = 1\% is reported, as in \cite{george-tifs-2019}.
Note however that in the following tables, both APCER and BPCER are reported \textit{on the test set}: 
the threshold reaching a BPCER of 1\% is selected \textit{a priori} on the validation set. 
As a consequence, applying the same threshold on the test set may lead to a slightly different BPCER. 

\subsection{Baselines \& Experimental Setup}
In this section, the baselines used for comparison to the proposed approaches are presented. Some of 
the implementation details are also provided.

\subsubsection{Baselines}

To assess our approach based on SWIR differences and CNNs in tackling the PAD problem, we compare its usage to different
baselines. First, we provide results for our own implementation - and adaptation - of the approach described in \cite{steiner-icb-2016}. 
The algorithm described in \cite{steiner-icb-2016} is actually a pixel-based classifier aiming at discriminating
skin from non-skin pixels. For this purpose, the authors used a so-called spectral signature as feature and a Support Vector Machine (SVM) as the classifier. 
The feature vector for a single pixel is the concatenation of 6 differences between different pre-selected SWIR wavelengths (935nm, 1060nm, 1300nm and 1550nm). 
In our work, this pixel-wise classifier has been adapted to perform presentation attack detection: 
the final score for a probe image is obtained by averaging the probabilities of skin-like pixels in the image. Note also that for training
such a model, and since annotations are not available at the pixel level, the following strategy has been applied:
the distribution of skin-like pixels has first been learned using a Gaussian Mixture Model. Then, a threshold on the likelihood of a pixel to be skin-like
has  been found considering both \bona and \textit{impersonation} attack examples in the training set. Finally, every pixels in all training images 
have been labelled as either skin or non-skin. A fraction\footnote{Since the total number of pixels in the training set is very large (100M+ examples), 
only a fraction of pixels in each image 
has been considered as training data for the SVM. Specifically, 1\% of pixels in each image has been retained, which yield a training set of 
approximately 351'428 positive, skin-like examples and 1'035'376 negative examples.} of these data have been used to train the SVM classifier.\\

We also provide results using CNNs acting on other image modalities (visible, infrared, thermal and depth), 
as proposed in previous works \cite{george-tifs-2019} \cite{george-icb-2019}. 
Finally, and for the sake of completeness, results are provided using the investigated architectures in conjunction with the SWIR differences
used in the context of fingerprint presentation attack detection \cite{tolosana-tifs-2019}. 

\subsubsection{Implementation Details}

Faces are first located in each of the 10 frames for each sequence using
an implementation of the MTCNN face detector \cite{zhang-spl-2016} in the visible spectrum. 
Facial landmarks are then detected and used to register face images in the different modalities.
Finally, face images are resized according to the different model requirements:
128x128 for the MC-CNN and 224x224 for the MC-PixBiS. Note that face images in all modalities but SWIR
are further preprocessed as in \cite{george-tifs-2019}. For the SVM baseline, a face size of 128x128 was
also used for consistency. The SVM has an RBF kernel with $\gamma = 0.1$. 
For the deep models, the MC-CNN is first initialized, in each channel, with a pre-trained Light CNN model \cite{wu-tifs-2018} before being trained for 50 epochs.
The MC-PixBiS is initialized with a DenseNet model pre-trained on ImageNet and is further trained for 30 epochs.
Note however that, at each epoch, a validation step is performed using the validation set: the model with the lowest
validation error is then further considered to assess the performance on the unseen test set. Other training parameters have been
set as in \cite{george-tifs-2019} and \cite{george-icb-2019} for MC-CNN and MC-PixBiS respectively. All experiments have been performed
using the bob toolbox \cite{anjos-icml-2017} and the code to reproduce all experiments presented in this paper is freely available
for download\footnote{\url{https://gitlab.idiap.ch/bob/bob.paper.pad_mccnns_swirdiff}}.

\subsection{Results}

In the following Tables, we present the performance of the baselines described above and for the two deep models used in 
conjunction with different combination of SWIR differences as input. For the baseline algorithms, note that $\Delta$SWIR\textsubscript{6} refers to the 6 SWIR differences
used in \cite{steiner-icb-2016}, GDIT stands for Grayscale, Depth, Infrared and Thermal, and color simply refers to RGB images. 
Two sets of SWIR differences have been used in conjunction with the two CNNs: $\Delta$SWIR\textsubscript{fp} stands for the (fixed) SWIR differences used in \cite{tolosana-tifs-2019} 
and $\Delta$SWIR\textsubscript{opt} refers to the best set of SWIR images differences found thanks to the SFFS algorithm (see Algorithm~\ref{algo:sffs}). Note
that the SFFS algorithm has been applied for each scenario. Results are presented
for the three scenarios described in Section~\ref{sec:protocols}.

\subsubsection{Generic Performance (Grand Test)}

\begin{table}[h!]
  \caption{BPCER, APCER and ACER [\%] on the test set of the Grand Test protocol.}
  \centering
  \begin{tabular}{ll|rrr}
    \textbf{Model} & \textbf{Input} & \textbf{BPCER} & \textbf{APCER} & \textbf{ACER}\\ 
    \toprule
                        SVM & $\Delta$SWIR\textsubscript{6}       & 2.7 & 62.6 & 32.6 \\
                        MC-CNN & GDIT \cite{george-tifs-2019}     & 0.0 & 59.8 & 29.9\\
                        PixBiS & color \cite{george-icb-2019}     & 0.1 & 15.7 & 7.9 \\
    \midrule
    \multirow{2}{*}{MC-CNN} & $\Delta$SWIR\textsubscript{fp}      & 0.0 & 10.0 & 5.0\\
                            & $\Delta$SWIR\textsubscript{opt}     & 6.0 & 7.2 &  6.6\\
    \midrule
    \multirow{2}{*}{MC-PixBiS} & $\Delta$SWIR\textsubscript{fp}   & 2.8 & 10.3 & 6.6\\
                               & $\Delta$SWIR\textsubscript{opt}  & \textbf{0.0} & \textbf{9.4} & \textbf{4.7}\\
    \bottomrule
  \end{tabular}
\label{tab:grand-test}
\end{table}

Table~\ref{tab:grand-test} shows the performance when all types of attacks are considered. 
As can be seen, the proposed approach combining MC-PixBiS with optimal SWIR differences found by the SFFS algorithm, $\Delta$SWIR\textsubscript{opt},
outperforms all other approaches, sometimes by a large margin. 
All approaches combining CNNs with SWIR differences perform better that deep models using
other modalities and the baseline SVM. This validates the assumption that deep models acting on SWIR differences
are effective for face PAD. Both architectures perform equally well when used in conjunction with
SWIR data, but the PixBiS using color 
information only performs better than using the MC-CNN acting on a several modalities including the visual spectrum. This suggest
that the binary pixel-wise supervision for face PAD introduced in \cite{george-icb-2019} is particularly efficient.

\subsubsection{Performance on Impersonation Attacks}

\begin{table}[h!]
  \caption{BPCER, APCER and ACER [\%] on the test set of the Impersonation protocol.}
  \centering
  \begin{tabular}{ll|rrr}
    \textbf{Model} & \textbf{Input} & \textbf{BPCER} & \textbf{APCER} & \textbf{ACER}\\ 
    \toprule
                        SVM & $\Delta$SWIR\textsubscript{6}      & 2.7 & 21.5& 12.1\\
                        MC-CNN & GDIT \cite{george-tifs-2019}    & 9.5 & 0.0 & 4.8\\
                        PixBiS & color \cite{george-icb-2019}    & 0.0 & 2.0 & 1.0\\
    \midrule
    \multirow{2}{*}{MC-CNN} & $\Delta$SWIR\textsubscript{fp}     & 2.0 & 0.0 & 1.0 \\
                            & $\Delta$SWIR\textsubscript{opt}    & \textbf{0.9} & \textbf{0.0} & \textbf{0.5}\\
    \midrule
    \multirow{2}{*}{MC-PixBiS} & $\Delta$SWIR\textsubscript{fp}  & 1.7 & 0.0 & 0.8 \\
                               & $\Delta$SWIR\textsubscript{opt} & 2.2 & 0.0 & 1.1  \\
    \bottomrule
  \end{tabular}
\label{tab:impersonation}
\end{table}

In the case of impersonation attacks, all approaches perform pretty well, and the best ones are close
to perfect performance, as can be seen in Table~\ref{tab:impersonation}. 

It should be noted that when using color information only (PixBiS + color), all \bona examples are correctly detected, as well as most of the attacks:
impersonation attacks usually exhibit different texture patterns and altered image quality as compared to \bona examples.
Consequently, it may not be necessary to add other sources of information. 

Nonetheless, all the approaches relying on SWIR difference images (i.e. the last 4 rows of Table~\ref{tab:impersonation}) achieve comparable or
better performance. Moreover, they are all capable of detecting all attacks, but at the cost of misclassifying some \bona attempts. Note however that the BPCER
remains very low, and this proves that SWIR information alone is at least as efficient as other modalities to detect impersonation attacks.
Also, these results suggest that SWIR image differences and color contain complementary information in the context of face PAD.

Finally, it should be noted that the SVM baseline generally performs worse than all the other approaches: this may be explained by the local, pixel-wise classification,
instead of a more "holistic" view, as performed by the CNN models.

\subsubsection{Performance on Obfuscation Attacks}
\label{sec:perf-obfuscation}

\begin{table}[h!]
  \caption{BPCER, APCER and ACER [\%] on the test set of the Obfuscation protocol.}
  \centering
  \begin{tabular}{ll|rrr}
    \textbf{Model} & \textbf{Input} & \textbf{BPCER} & \textbf{APCER} & \textbf{ACER}\\ 
    \toprule
                        SVM & $\Delta$SWIR\textsubscript{6}      & 2.7 & 99.8 & 51.2\\
                        MC-CNN & GDIT \cite{george-tifs-2019}    & 0.3 & 47.1 & 23.7\\
                        PixBiS & color \cite{george-icb-2019}    & \textbf{0.1} & \textbf{21.0} & \textbf{10.5}\\
    \midrule
    \multirow{2}{*}{MC-CNN} & $\Delta$SWIR\textsubscript{fp}     & 1.9 & 27.7 & 14.8\\
                            & $\Delta$SWIR\textsubscript{opt}    & 6.4 & 28.6 & 17.5 \\
    \midrule
    \multirow{2}{*}{MC-PixBiS} & $\Delta$SWIR\textsubscript{fp}  & 0.0 & 27.4 & 13.7\\
                               & $\Delta$SWIR\textsubscript{opt} & 0.0 & 23.1 & 11.5\\
    \bottomrule
  \end{tabular}
\label{tab:obfuscation}
\end{table}

As evidenced by the error rates reported in Table~\ref{tab:obfuscation}, obfuscation attacks are generally harder to detect than impersonation attacks.
This makes sense, since they are more subtle and usually only affect a portion of the face, as opposed to impersonation attacks, where the whole face is covered. 
Here, the best performance is obtained with the PixBiS model using color information only. This was not expected, since in the more generic "Grand Test" case,
the performance obtained with SWIR image differences is generally better. This led us to have a closer look on the results, and consequently, a breakdown per attack type
is presented in Table~\ref{tab:obfuscation-breakdown}.

\begin{table}[h]
\centering
  \caption{APCER [\%] for different attacks on the test set of the Obfuscation protocol.}
  \begin{tabular}{lcc}
    \toprule
    & PixBiS + color & MC-PixBiS + $\Delta$SWIR\textsubscript{opt} \\
    \midrule
    Glasses   & 69.3  & 0.6\\
    Makeup    & 7.7   & 13.8\\
    Tattoo    & 0.0   & 95.7\\
    Wig       & 95.2  & 94.7\\
    \bottomrule
  \end{tabular}
  \label{tab:obfuscation-breakdown}
\end{table}

Table~\ref{tab:obfuscation-breakdown} offers an interesting insight and clearly shows the differences between the two approaches. The model relying on 
color information is good at detecting Makeup and Tattoo whereas it fails on Glasses and Wigs. On the other hand, MC-PixBiS + $\Delta$SWIR\textsubscript{opt}
performs very well on Glasses attacks, but very poorly on Tattoos. These results are not surprising: tattoos do not actually appear in the SWIR spectrum, as opposed to 
glasses (thanks to the different material). Again, this suggest that these two sources of information complement each other.
Note finally that in this case, SVM performs very poorly since it is pixel-based, and that in most cases, the number of skin-like pixels are greater than non-skin pixels. 
Consequently, this approach is not suitable for generic face PAD \textit{only} and should be coupled with a face recognition system (as proposed in \cite{steiner-icb-2016}).

\subsection{Discussion}

Several observations can be made from the results presented above. First and foremost, it was shown that the conjunction of SWIR differences and CNNs
is indeed successful in face PAD and achieve relatively low error rates. This is an interesting result for several reasons. Firstly, it shows that
SWIR information should be considered at the global image level, as it is the case with CNNs, rather than considering it at the pixel level (as in the SVM case).
This is especially true for obfuscation attacks, where the number of altered pixels are not known, and vary (as opposed to impersonation attacks, where the whole image has been altered).
Secondly, while the PixBiS + color model acting performs well, using SWIR data yields comparable and even better performance
across all considered scenarios. As shown in Table~\ref{tab:obfuscation-breakdown}, one can see that these two modalities are
clearly complementary to each other and this opens new directions for future research.

Table~\ref{tab:opt-swirdiff} shows the optimal
set of differences (see Equation~\ref{eq:swir-diff}) for each scenario. As it can be seen, the selected
differences are not the same depending on the type of attacks. It shows that applying a feature selection algorithm instead of 
using a fixed set of pre-defined differences is relevant, since optimal features are task-dependant. 

\begin{table}[!h]
  \caption{Optimal SWIR differences for MC-CNN in each scenarios. $s_1$ and $s_2$ refer to the SWIR wavelengths in Equation~\ref{eq:swir-diff}.}
  \centering
  \begin{tabular}{cc|cc|cc}
    \toprule
    \multicolumn{2}{c}{Grand Test} & \multicolumn{2}{c}{Impersonation} &  \multicolumn{2}{c}{Obfuscation}\\
    \cmidrule(lr){1-2}
    \cmidrule(lr){3-4}
    \cmidrule(lr){5-6}
    $s_1$ & $s_2$ & $s_1$ & $s_2$ & $s_1$ & $s_2$\\
    \midrule
    1550 & 1200 & 1550 & 1200 & 1450 & 1200  \\
    1450 & 1200 & 1450 & 1200 & 1550 & 1050  \\
    1200 & 1550 &  -   & -    & 1200 & 1550            \\
    940 & 1550 &  -   & -     & 1200 & 1450            \\
    940 & 1650 &  -   & -     & 1650 & 1050            \\
    - & - &  -   & -          & 1450 & 1550            \\
    \bottomrule
  \end{tabular}
  \label{tab:opt-swirdiff}
\end{table}

Several additional observations can be made from this table. Firstly, only a few differences seem to
be relevant for face PAD: remember that the SFFS algorithm considered an initial pool of 42 SWIR differences as input. Secondly, less 
features are needed when the variability of attacks is limited. Indeed, for impersonation attacks,
only 2 SWIR differences are used to reach optimal performance. When the set of different attacks is enlarged, as it is the
case in the last scenario, more features are needed. Note also that depending
on the type of attacks, optimal features are not the same. This again advocates for a mechanism to select
relevant features, depending on the scenario. Finally, it is interesting to see that in all cases, considered 
wavelengths fall on one or the other side of 1430nm. This is not surprising, since water
absorption peaks at around 1430nm and hence skin appears very dark at this wavelength.

\subsection{Cross-database experiment}

Cross-database experiments have been conducted to gauge the generalization ability of deep CNNs using SWIR data. As mentioned in Section~\ref{sec:introduction},
the \textit{only} database containing \bona face images and spoofing attacks imaged in both color and SWIR domain is the BRSU database \cite{steiner-sensors-2015}. 
As compared to the proposed \dbname database, BRSU only contains images at 4 different SWIR wavelengths: 935nm, 1060nm, 1300nm and 1550nm. 
Besides, this database only contains 276 \textit{frontal} face images (192 \bona and 84 attacks), and it is thus not possible
to train the proposed models with so little data. Consequently, models were first trained on \dbname and then evaluated on the 276 images from BRSU. 
More specifically, the SFFS algorithm was applied to find optimal SWIR differences, but only considering differences available within BRSU. \\

\begin{figure}[h!]
  \centering
  \includegraphics[height=4cm]{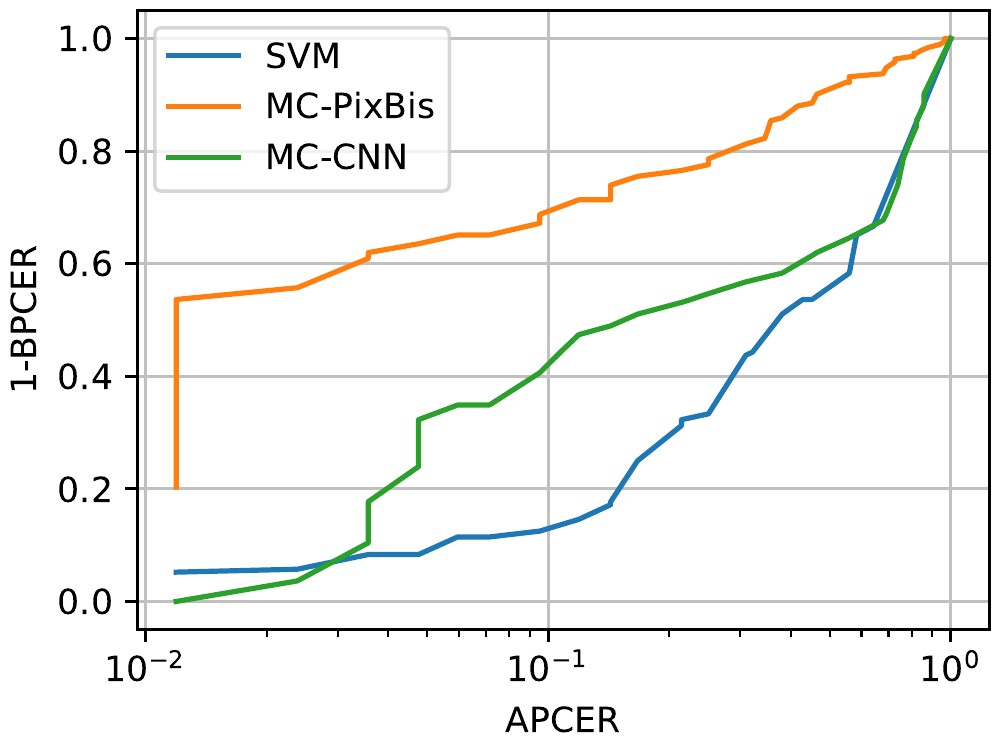}
  \caption{ROC curves for the SVM baseline and both CNN models with optimal SWIR differences on the BRSU database.}
  \label{fig:roc-brsu}
\end{figure}

Since BRSU contains few data, no subset has been used for validation. As a consequence, one cannot set a decision threshold \textit{a priori}. Results
are hence presented as ROC curves. As can be seen on Figure~\ref{fig:roc-brsu}, performance is far from being satisfactory on this database for both the MC-CNN model and 
for the SVM baseline. MC-PixBiS, although overall better, does not generalizes so well since it reaches an Equal Error Rate (EER, when the threshold
is selected such that BPCER = APCER) of 22.8\%. 

\begin{figure}[!h]
  \centering
  \includegraphics[height=7cm]{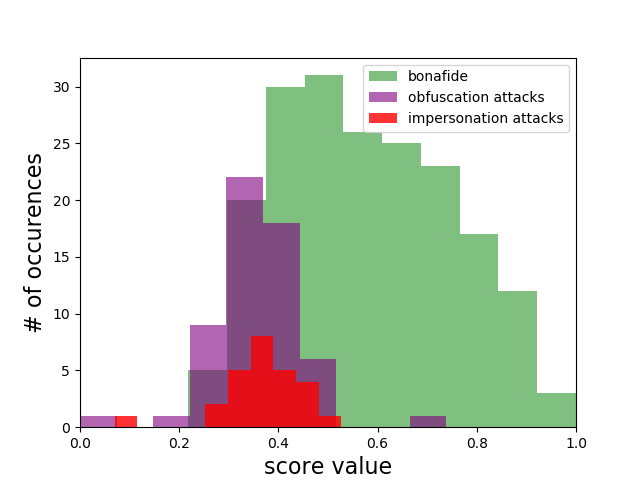}
  \caption{Scores distribution (given by MC-PixBiS) on the BRSU database.}
  \label{fig:brsu-distribution}
\end{figure}

To go one step further, scores distribution on \bona images, obfuscation and impersonation attacks are presented in Figure~\ref{fig:brsu-distribution}. This clearly
shows that the main issue occurs with \bona data. Indeed, most of the scores for both impersonation and obfuscation attacks are relatively low (i.e. $<$ 0.5), but 
scores obtained on \bona examples are more spread, with a median of 0.56. Tentative explanations for the distribution of \bona scores include
i) SWIR wavelengths present in the BRSU database may not be the most suited for our models and ii) the differences between \bona training data from the \dbname database 
and testing data from the BRSU database. 

\section{Conclusion}
\label{sec:conclusion}
In this contribution, two recent models for face PAD based on deep convolutional neural networks have 
been investigated in conjunction with SWIR image differences. They have been compared to baselines using 
either color or a combination of other modalities (visible, infrared, depth and thermal imaging), 
as well as to the adaptation of a previous approach acting on SWIR data. For this purpose, a new database for face presentation attack detection has 
been introduced. \textit{Bonafide} attempts and presentation attacks have been recorded in 
several modalities, including the short-wave infrared spectrum, which makes it particularly interesting
to develop new approaches leveraging SWIR imaging properties. 
Besides, this database contains a large variety of attacks, that can be split into two
categories: impersonation and obfuscation. Impersonation attacks consists of various print, replay
and mask attacks while obfuscation attacks comprise different variations of glasses, wigs, makeup
and tattoos.

Experimental results show that the performance of investigated CNN models with carefully selected SWIR differences outperform baselines
when a large variety of attacks is considered. 
Furthermore, combining deep models for face PAD together with SWIR differences 
allows to almost perfectly detect all impersonation attacks while maintaining a very low BPCER. 
However, it should be noted that attacks aiming at hiding one's identity - as opposed to impersonating
someone else - are harder to detect: this suggests interesting directions for future research. Finally, the
generalization ability of the different models using SWIR data has been assessed on a cross-database experiment using 
the \textit{only} other publicly available PAD database containing SWIR data. 
In this case, a noticeable difference is observed on \bona attempts: when trained and evaluated on different
data, proposed models do not generalize well. This can be explained by the usage of different wavelengths in
the SWIR spectrum, or this can be due to the difference in image quality between the two databases.

Note finally that the proposed database, as well as the code and instructions to reproduce presented experiments
have been made freely available to download for research purposes. This will certainly foster further
research efforts on face presentation attack detection using data from several image modalities.

\section*{Acknowledgments}
Part of this research is based upon work supported by the Office of the Director of National Intelligence (ODNI), Intelligence Advanced Research Projects Activity (IARPA), via IARPA R\&D Contract No. 2017-17020200005. The views and conclusions contained herein are those of the authors and should not be interpreted as necessarily representing the official policies or endorsements, either expressed or implied, of the ODNI, IARPA, or the U.S. Government. The U.S. Government is authorized to reproduce and distribute reprints for Governmental purposes notwithstanding any copyright annotation thereon.

\ifCLASSOPTIONcaptionsoff
  \newpage
\fi



\bibliography{references}
\bibliographystyle{IEEEtran}

\begin{IEEEbiography}
  [{\includegraphics[width=1in,height=1.25in,clip,keepaspectratio]{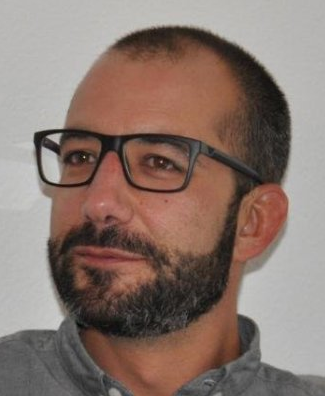}}]{Guillaume Heusch}
received a MSc. in Communication Systems and a Ph.D. in Electrical Engineering, 
both from Ecole Polytechnique F{\'e}d{\'e}rale de Lausanne (EPFL) in 2005 and 2010 respectively. 
He then spent several years working as a computer vision research engineer in various industries. 
He is now a research associate at Idiap Research Institute. His current research interests are computer vision, machine learning 
and, on a broader perspective, the extraction of meaningful information from raw data.
\end{IEEEbiography}

\begin{IEEEbiography}
  [{\includegraphics[width=1in,height=1.25in,clip,keepaspectratio]{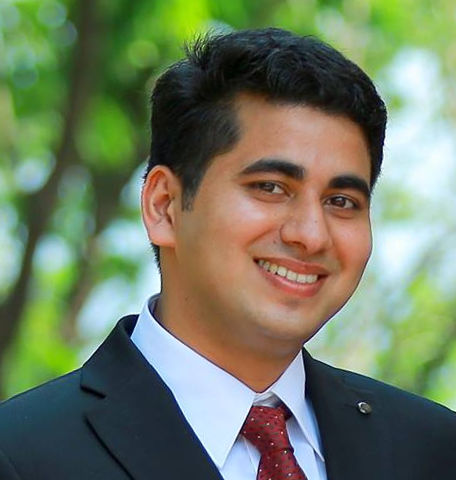}}]{Anjith George}

has received his Ph.D. and M-Tech degree from the Department of Electrical Engineering, Indian Institute of Technology (IIT) Kharagpur, India in 2012 and 2018 respectively. After Ph.D, he worked in Samsung Research Institute as a machine learning researcher. Currently, he is a post-doctoral researcher in the biometric security and privacy group at Idiap Research Institute, focusing on developing face presentation attack detection algorithms. His research interests are real-time signal and image processing, embedded systems, computer vision, machine learning with a special focus on Biometrics.
\end{IEEEbiography}

\begin{IEEEbiography}
  [{\includegraphics[width=1in,height=1.25in,clip,keepaspectratio]{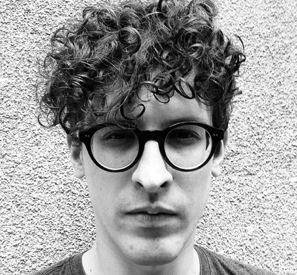}}]{David Geissb\"uhler}

 is a researcher at the Biometrics Security and Privacy (BSP) group at the Idiap Research Institute (CH) and conducts research on multispectral sensors. He obtained a Ph.D. in High-Energy Theoretical Physics from the University of Bern (Switzerland) for his research on String Theories, Duality and AdS-CFT correspondence. After his thesis, he joined consecutively the ACCES and Powder Technology (LTP) groups at EPFL, working on Material Science, Numerical Modeling and Scientific Visualisation.
\end{IEEEbiography}

\begin{IEEEbiography}
  [{\includegraphics[width=1in,height=1.25in,clip,keepaspectratio]{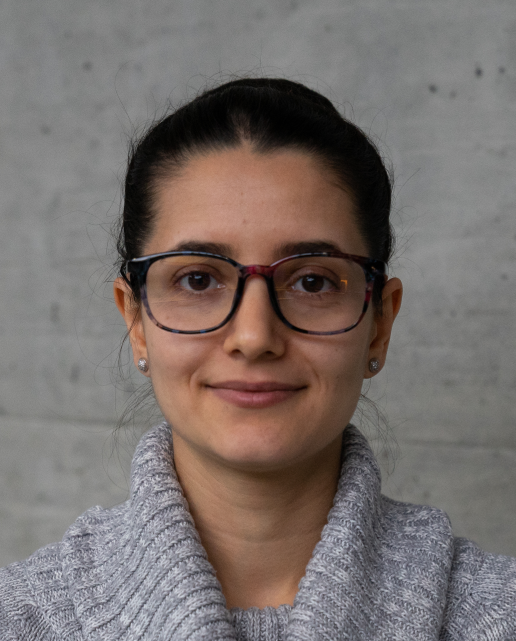}}]{Zohreh Mostaani}

obtained the B.Sc. in Electrical Engineering from University of Tehran, Iran, and M.Sc. in Electrical and Electronics Engineering from Ozyegin University, Turkey. She worked as a Research and Development Engineer in the Biometrics Security and Privacy group at Idiap Research Institute, Switzerland. She is currently a PhD student in the Speech and Audio Processing group at Idiap. Her research interests are Machine learning, Speech processing, and Biometrics.
\end{IEEEbiography}

\begin{IEEEbiography}
 [{\includegraphics[width=1in,height=1.25in,clip,keepaspectratio]{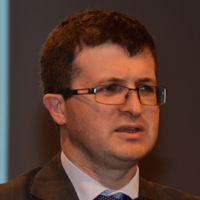}}]{S{\'e}bastien Marcel}
received the Ph.D. degree in signal processing from Universit\'{e} de Rennes I, Rennes, France, in 2000 at CNET, the Research Center of France Telecom (now Orange Labs). He is currently interested in pattern recognition and machine learning with a focus on biometrics security. He is a Senior Researcher at the Idiap Research Institute (CH), where he heads a research team and conducts research on face recognition, speaker recognition, vein recognition, and presentation attack detection (anti-spoofing). He is a Lecturer at the Ecole Polytechnique F\'{e}d\'{e}rale de Lausanne (EPFL) where he teaches a course on ``Fundamentals in Statistical Pattern Recognition.'' He is an Associate Editor of IEEE Signal Processing Letters. He has also served as Associate Editor of IEEE Transactions on Information Forensics and Security, co-editor of the ``Handbook of Biometric Anti-Spoofing,'' Guest Editor of the IEEE Transactions on Information Forensics and Security Special Issue on ``Biometric Spoofing and Countermeasures,'' and co-editor of the IEEE Signal Processing Magazine Special Issue on ``Biometric Security and Privacy.'' He was the Principal Investigator of international research projects including MOBIO (EU FP7 Mobile Biometry), TABULA RASA (EU FP7 Trusted Biometrics under Spoofing Attacks), and BEAT (EU FP7 Biometrics Evaluation and Testing). 
\end{IEEEbiography}

\end{document}